\documentclass[conference]{IEEEtran}
\usepackage{cite}
\usepackage{amsmath,amssymb,amsfonts}
\usepackage{algorithmic}
\usepackage{graphicx}
\usepackage{amsmath}
\graphicspath{ {images/} }
\usepackage{textcomp}
\usepackage{xcolor}
\usepackage{caption}
\def\BibTeX{{\rm B\kern-.05em{\sc i\kern-.025em b}\kern-.08em
    T\kern-.1667em\lower.7ex\hbox{E}\kern-.125emX}}
\begin{document}

\title{Recurrent Neural Networks for video object detection}

\author{\IEEEauthorblockN{Bin Qasim Ahmad}
\IEEEauthorblockA{\textit{Technical University of Munich} \\
\textit{Department of Informatics}\\
Munich, Germany \\
ahmad.qasim@tum.de }
\and
\IEEEauthorblockN{Pettirsch Arnd}
\IEEEauthorblockA{\textit{Technical University of Munich} \\
\textit{Department of Informatics}\\
Munich, Germany \\
a.pettirsch@outlook.de}
}

\maketitle

\begin{abstract}
There is lots of scientific work about object detection in images. For many applications like for example autonomous driving the actual data on which classification has to be done are videos. This work compares different methods, especially those which use Recurrent Neural Networks to detect objects in videos. We differ between feature-based methods, which feed feature maps of different frames into the recurrent units, box-level methods, which feed bounding boxes with class probabilities into the recurrent units and methods which use flow networks. This study indicates common outcomes of the compared methods like the benefit of including the temporal context into object detection and states conclusions and guidelines for video object detection networks. 
\end{abstract}


\section{Introduction}

\subsection{Image and Video Object Detection in general}
\subsubsection{History of image object detection} 
Before the advent of Deep Learning, image detection was carried out using classic Machine Learning methods like Discriminative Models e.g. Logistic Regression, Support Vector Machines etc. or Generative Models e.g. Gaussian Mixture Models etc. Bag of Visual Words (BOVW) was another technique that was used for the same purpose. The use of Deep Learning methods for image detection was restrained by two factors, first of all the dearth of large amounts of labeled data and secondly, the computational power. In 2009, Jia Deng et. al. presented a dataset named ImageNet\cite{b13} containing millions of images labeled into different categories. A breakthrough paper published in the same year by Rajat Raina\cite{b46} et. al. stipulated that using GPUs for deep learning can provide the much-needed computational power. Hence, the door to using deep learning for image classification and detection were open and the first landmark paper in this domain was published in 2012 by Alex Krizhevsky\cite{b14} et. al. \newline

\subsubsection{Types of image object detectors} 
Image object detectors can be categorized into two types, single stage and 2-stage image object detectors. A two-stage pipeline firstly generates region proposals, which are then classified and refined\cite{b17} while a single-stage method, which is often more efficient but less accurate, directly regresses on bounding boxes and classes\cite{b18}\cite{b19}. \newline

\subsubsection{Why is video object detection harder?} 
	\begin{itemize}
		\item Large number of individual frames within the videos
		\item The occurence of motion blur between different frames
		\item The quality of video datasets are often not ideal
		\item The objects to be detected can get partially occuluded
		\item The objects can have unconventional poses
	\end{itemize}

\subsection{Recurrent Neural Networks in general}
\subsubsection{Delimitation to non-recurrent Neural Networks} 
Non-recurrent Neural Networks process on single inputs for example, a single image. Recurrent Neural Networks process on sequences of data, e.g.: multiple video frames. Recurrent Neural Network's core concept to enable the sequence processing is parameter sharing across different parts of a model. Parameter sharing can be reached by cycles in the architecture. \cite{b11} \newline

\subsubsection{Common Types of Recurent Neural Networks}
Most of the papers which are described in this work use two common Recurrent Neural Network approaches. The first one are LSTMs, first mentioned in \cite{b18} in 1997. The key of LSTMs are different gates (forget gate, external input gate, output gate) to control the cell state and the hidden state of the LSTM \cite{b11}.  \newline

The second type of Recurrent Neural Networks are Gated Recurrent Units (GRUs). The main difference to LSTMs is, that those GRUs consist out of a single gated unit which can simultaneously control the forgetting factor and decide to update the inner cell state. \cite{b11} 

\section{Feature-based Video Object Detection}

First, we want to introduce feature-based Video Object Detection methods. As defined, for example, in [1-2] feature-based Video Object Detection methods fuse detectors which integrate features from multiple frames into their video detection. In most papers considered in this work this integration is done by recurrent units.  

\subsection{Recurrent Multi-frame Single Shot Detector for Video Object Detection [1]}
Broad, Jones and Lee have in \cite{b1} and \cite{b12} the idea to design a multi-frame video object detector by inserting a fusion-layer into a classical Single Shot Detector (SSD) meta-architecture. Based on this idea they research in two main fields: On the one hand, they investigate different fusion methods and on the other hand they try several SSD meta-architectures [20-2.1.1; 20-3.6.1]. They test their approaches on the KITTI dataset \cite{b21} for autonomous driving and their model improves upon a state-of-the-art SSD model by 2.7 \% mAP. Finally they evaluate their best approach on Caltech Pedestrians Dataset \cite{b22} and find similar improvements of 5 \% in maP.   \newline

SSDs in general consist of two main components. First, an so called Feature Extractor, which gets as an input an image and outputs feature-maps. The other component is an Detection head which creates bounding boxes and class probabilities out of the feature maps. [1-3] Broad, Jones and Lee have the idea to insert a fusion layer in between those two components. As fusion techniques they test simple element-wise operations (e.g.: add or max), simple concatenations of features maps and recurrent layers [1-3]. Their experiments show that the recurrent layers perform best, because they add in contrast to the other two method new parameters to learn temporal context to the network. In addition, they observe that recurrent units do not slow down the computational speed significantly (only 5 fps slower then the baseline SSD) [1-4.1]. As a recurrent unit they use GRU because they observe that the results were similar to the results when they use LSTMs but the GRUs are faster to train [1-3.1.1]. Their final architecture is shown in Figure 1. 

\begin{figure} [h]
\includegraphics[width=\columnwidth]{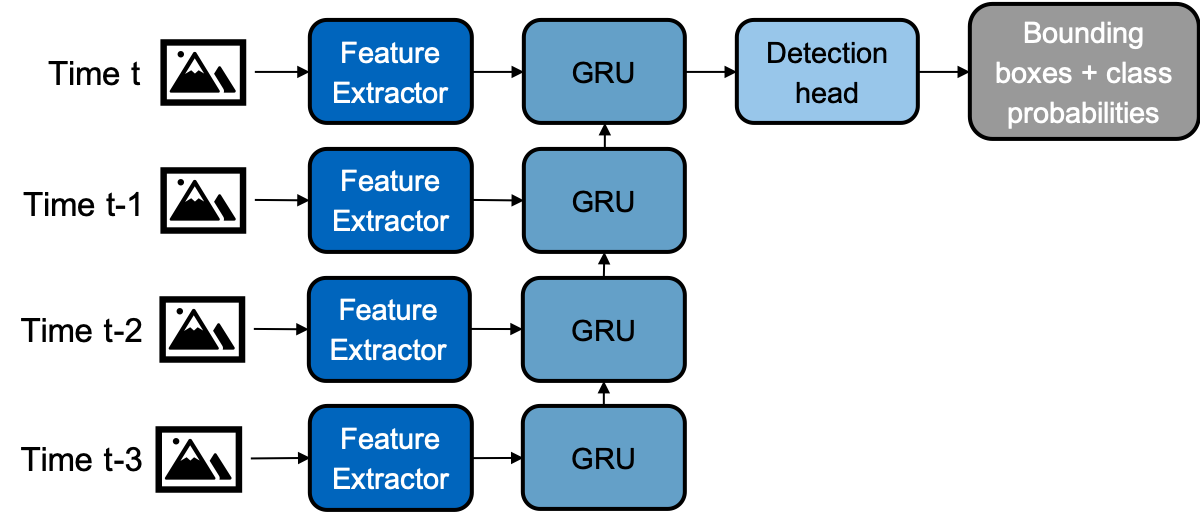}
\caption{Architecture: Recurent Multi-frame single shot detector [1]. The Feature Extractor generates feature maps for four frames and feeds those feature maps into the GRUs (Gated Recurrent Units). The Detection Head uses the final feature maps which are created by the GRU with respect to the temporal context to create Bounding Boxes and Class probabilities}
\end{figure}

In addition Broad, Jones and Lee test different types of SSDs as a baseline for their architecture. For all baseline SSDs the mAP was higher in comparison to the non-recurrent models. The mAP increase by 1.4 to 4.4 percent on KITTI dataset.  The best mAP is achieved using SqueezeDet+ \cite{b23} as a baseline SSD network [1-4.2].  \newline

Finally Broad, Jones and Lee use the Caltech Pedestrians dataset to expolore the effect of the number of prior frames and the frame-rate. They compare the single-frame SSD model, with the RMf-SSD using the prior frames and the RMf-SSD model using the frames t-2, t-4 and t-6. The improvement was 3 percentage points higher by using the frames t-2, t-4 and t-6 in comparison to use the prior 3 frames [1-4.4]. \newline  

There is only a little information on the training of the Recurrent Multi-frame Single Shot Detector. They use the SequeezeDet Training strategies [23-3.3] and a pre-trained version of the baseline SSD. Finally, they use the SequezzeDet fine tuning strategy to train the whole network afterwards [12-2]. \newline

Overall Broad, Jones and Lee show three main things. First they show that including the temporal context improves single frame SSD methods. Moreover they show that Recurrent Units are a good approach to add the temporal context and finally they point that long time context (frames t-2, t-4, t-6) seems to be more important then short-term temporal context. Finally they reach an mAP of 83 \% on KITTI dataset [1-4.2] and a miss-rate of 29 \% on the Caltech Pedestrians dataset [1-4.1]. \newline

This architecture is comparable to the architectures described in 2-B and 2-E. Main differences are that 2-B uses more then one recurrent unit and 2-E different feature extractors for different frames. But all of them feed feature maps of different frames into recurrent units. 
 
\subsection{Mobile Video Object Detection with Temporally Aware Feature Maps [2]}
Liu and Zhu have in \cite{b2} the goal to design a video object detection architecture, which can run real-time (15 fps) on low-powered mobile and embedded devices. The key of their method is to combine convolutional layers with convolutional LSTMs. They investigate on the benefit from adding an LSTM into the baseline SSD, different types of Recurrent Layers (LSTM, GRU and bottleneck LSTMs), different dimensions of their bottleneck LSTMs and different LSTM placement strategies (Single LSTM placement and multiple LSTM placement). They test their model on Imagenet VID 2015 dataset \cite{b35} and reached a mAP of 54.4 \% while performing on 15 fps.  \newline

In the beginning they simply add one LSTM to their baseline SSDs architecture - MobileNet \cite{b24}. They observe that adding the LSTM improves the mAP in comparison to their baseline SSD architecture. Moreover, they investigate that the greatest improve is by adding the LSTM after the 13th convolutional Layer of the SSD. [2-4.2] \newline

Afterwards, they compare LSTMs, GRUs and Bottleneck LSTMs by placing them after the 13th convolutional layer. Bottleneck-LSTMs have been designed by Liu and Zhu to increase the efficiency of LSTMs. They use convolutions with ReLU as activation function and feed Bottleneck feature map with less input channel then the original feature map into the LST. They come to the conclusion that Bottleneck-LSTMs are more effective the LSTMs and in the case of a convolutional kernel greater 1x1 even more effective than GRUs while attaining comparable perfomance. [2-3.4; 2-4.2]\newline

In addition to the bottleneck LSTMs Liu and Zhu extend their network with multipliers. They use $\alpha_{base}$, $\alpha_{ssd}$ and $\alpha_{lstm}$ as multiplier to scale the channel dimension of each layer. They find out that the accuracy of the model remains near constant up to $\alpha_{ssd}= 0.25 \alpha$. For the other multipliers they use the values: $\alpha_{base} = \alpha$ and $\alpha_{ssd} = 0.5 \alpha$. [2-3.3; 2-4.2] \newline

As shown in Fig. 2 the final model uses LSTMs after each convolutional Layer, because Liu and Zhu observe that there is a slight performance improve by adding LSTMs after every feature map and nearly no change in computational cost [2-4.2].

\begin{figure} [h]
\includegraphics[width=\columnwidth]{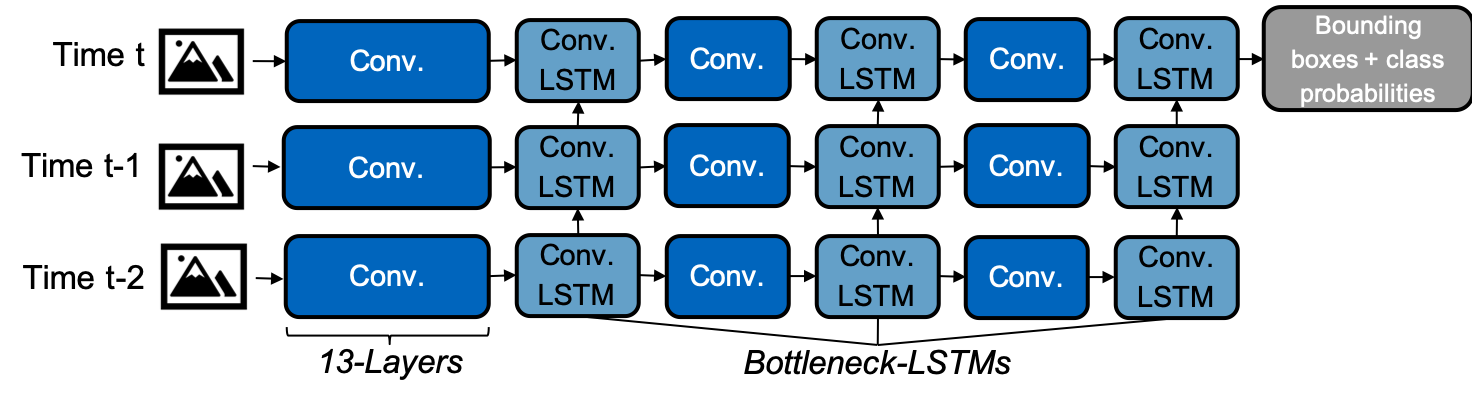}
\caption{Architecture Mobile Video Object Detection with temporally aware feature maps \cite{b2}. First 13 Convolutional Layeer create Feature Maps. Those Feature Maps are fed into the convolutional Bottleneck-LSTMs (faster LSTMs which use convolutions). Afterwards Convolutional-Layer create new Feature Maps and feed them again into Bottleneck-LSTMs.}
\end{figure}

On the training strategy and the loss function Liu and Zhu do not provide any information. \newline

Liu and Zhu confirm the observation from \cite{b1} that adding temporal context information with the help recurrent units improves the detection quality of the baseline SSDs. Moreover they show that architectures using recurrent units can still perform in real-time (15 fps) even on mobile devices. In comparison to the Mobilnet-SSD model they improve the mAP by 4.1 percentage and perform faster on Pixel 2 Phone. Overall they reach an 54.4 \% on Imagenet VID. \newline

This architecture can be especially compared to the model described in "Looking Fast and Slow" because both of them are processing on a mobile device and are feature-based architectures. But "Looking Fast and Slow" differs because it does not treat every frame in the same way and proceeds with different features extractors on different frames. 

\subsection{Feature Selective Small Object Detection via Knowledge-based recurrent attentive neural networks \cite{b6}}

[I] The aim of this paper is develop a video object detector for the purpose of autonomous driving. The network developed in this paper is termed as Knowledge-Based Recurrent Attentive Neural Network (KB-RANN). \newline
The main contributions of the authors, Kai Yi et. al., are:
\begin{itemize}
  \item An attention mechanism that works like human cognition. The attention mechanism can detect the salient features, which are important for the object detection problem and propagate them forward. This attention mechanism is based on previous work by Ashish Vaswani et. al. \cite{b30}. 
  \item A domain and intuitive knowledge module which can use the knowledge about traffic signals to produce feature maps.
  \item A model which has good ability for transfering knowledge. The authors obtained good results by evaluating the KB-RANN model on BTSD dataset \cite{b29}, which was trained on KITTI dataset \cite{b21}.
\end{itemize}

The KB-RANN model is tested on KITTI and MS COCO17 dataset \cite{b31}, achieving 0.813 mAP and 0.578 mAP repectively, on all classes of those datasets. The model performs better then several popular objection detection algorithms like Faster R-CNN \cite{b15}, RetinaNet \cite{b32} and SqueezeDet \cite{b33}.
The authors successfully compressed and accelerated their proposed model and deployed it to their own self-developed autonomous car. \newline

\begin{figure}[h]
\includegraphics[width=\columnwidth]{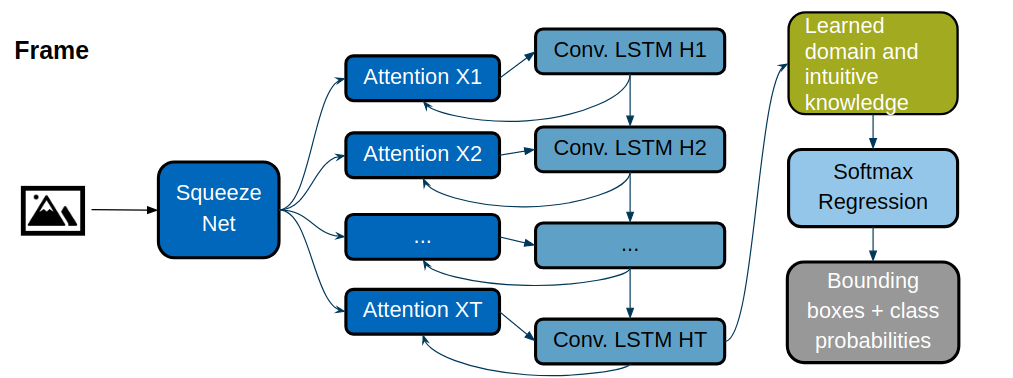}
\caption{Architecture of the model proposed in Feature Selective Small Object Detection via Knowledge-based recurrent attentive neural networks}
\end{figure}

[III-A] The authors use SqueezeNet\cite{b34} for feature extractor because SqueezeNet can provide AlexNet\cite{b14} level of accuracy but with 50 times less parameters. Although there are better performing models like VGGNet\cite{b45} and ResNet\cite{b40}, but these models are computationally expensive. They make a few changes in the SqueezeNet architecture, which include changing the kernel size and also fine-tuning the backend by adding two fire modules at the end. This modified SqueezeNet architecture is termed as, SqueezeNet+. \newline

[III-B] Attention mechanism is used to find the saliency maps from the deep feature maps obtained from SqueezeNet. Attention module obtains the input tensor X and outputs the saliency map $\tilde{X\textsubscript{t}}$. \newline

[III-B] LSTM \cite{b18} is used as a memory mechanism in order to find long term dependencies between different frames and the characteristics of Attention and LSTM are exploited together by fusing them with each other, this fused module is termed as RANN. As saliency feature extraction is applied on the original deep feature maps, it is possible that some of the information is lost. Multiple RANNs are cascaded together into a chain to minimize the effect of feature information loss. Moreover, the output H\textsubscript{t} of each LSTM is concatenated with the original feature map and used as an input for the next attention module to refine the saliency feature extraction process. \newline

[III-C] Lasty, The authors also add a domain and intuitive knowledge module to the KB-RANN architecture. It is assumed that traffic signs detection is one of the most important aspect of autonomous driving. The major focus of attention of a driver is at the center of vision and the traffic lights are always located at a bias from that central region of peoples' gazes. With these assumptions in place, domain knowledge about traffic signals is learned from the data itself by constraining the distribution to a 2D Gaussian function and learning the mean and covariance matrices from the data. A reverse Gaussian distribution is computed from the learned Gaussian distribution. The feature maps obtained from this reverse Gaussian distribution are concatenated with the feature maps from RANN. \newline

[IV-A,B,C]For training and evaluation of the model, the authors use the KB-RANN model and MS COCO17 dataset. They compare the results with other popular object detection models. On the KITTI dataset the KB-RANN achieves the mAP of 0.813, compared to the 0.763 of SqueezeDet, 0.702 of Faster R-CNN and 0.601 of RetinaNet. The frames-per-second (FPS) at which KB-RANN operates are higher then the compared models. In order to signify the accuracy gain from the attention mechanism and the knowledge module, authors also train and evaluate different architectures, namely KB-RCNN in which the attention mechanism is replaced with convulational layers and RANN, which does not have knowledge-based module. KB-RANN achieves better accuracy then KB-RCNN and RANN on KITTI and MS COCO17 dataset, although the FPS achieved by KB-RCNN due to its recurrent nature are higher. Lastly, a KB-RANN model with parameters trained on the KITTI dataset is also tested on BTSD dataset to demonstrate the knowledge transfer capabilities.

\subsection{Looking fast and slow: memory-guided mobile video object detection \cite{b7}}

[1] Mason Liu et. al. aim to, replicate the capability of a human visual system of obtaining the "gist" of a scene. Using this sparse information, and amalgamating it with the more thorough information, a human visual system can effectively detect objects in its field of vision. In light of this, the main contributions of this paper are:
\begin{itemize}
	\item The introduction of multiple feature extractors. Some of those feature extractors are very light-weight but can provide a "gist" of the frame, while others can provide more accurate representation of a frame but at the cost of performance.
	\item A memory module, which can fuse the outputs of those feature extractors.
	\item An adaptive interleaving policy which uses reinforcement learning to find to decide, which feature extractor should be executed.
	\item The capability of executing multiple feature extractors asynchronously. So that, the light-weight feature extractors don't have to wait for the more expensive feature extractors.
\end{itemize}
The model is evaluated on the ImageNet VID 2015 dataset \cite{b35}, which is concatenated with extra data from the ImageNet DET \cite{b35} and MS COCO17 datasets \cite{b31}. The model manages to achieve a mAP of 0.593 at 72.3 FPS on a Pixel3 mobile device. \newline

\begin{figure}[h]
\includegraphics[width=\columnwidth]{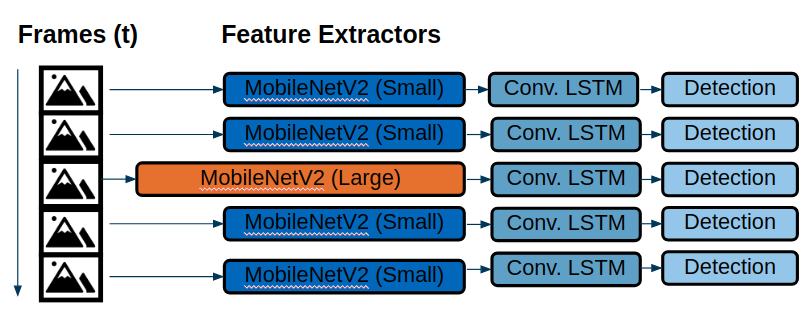}
\caption{The KB-RANN architecture. The frames are passed through the feature extractors to obtain, feature maps. The blue MobileNetV2 feature extractors are the light-weight ones while the red one is computationally more expensive.}
\end{figure}

[3-3.1]Let \textit{m} be the number of feature extractors. For the implementation of this paper, the authors constraint $\textit{m}=2$, a d let f\textsubscript{1} be the feature extractor which is computationally more expensive while f\textsubscript{2} be the feature extractor which is cheaper. Both feature extractors use MobileNetV2 architecture \cite{b36}. f\textsubscript{1} uses a depth of 1.4 with $320*320$ input frames resolution while f\textsubscript{2} uses a depth of 0.35 with a lower $160*160$ input frames resolution. \newline

[3-3.2] A modified LSTM \cite{b18} is used as a memory mechanism to preserve long term dependencies. The modifications in the LSTM are responsible for making the memory mechanism faster and also better at preserving long term dependencies. For faster memory mechanism, the authors make three modifications. They introduce bottlenecking and add a skip connection between the bottleneck and the output. Lastly, the LSTM states are grouped together and convulations are applied on each group seperately and the resultant states are then concatenated together to obtain the final states. The grouped convulations provide a speed-up. In order to improve the preservation of long term dependencies by the LSTM, the LSTM states are only updated when f\textsubscript{1} is run and not the f\textsubscript{2}, as feature maps obtained from f\textsubscript{2} are not of a higher quality compared to f\textsubscript{1} and this can result in loss of important state information in the LSTM. \newline
SSD-style \cite{b17} detection is applied on refined feature maps obtained from the LSTM for classification and obtaining bounding boxes. \newline

\begin{figure}[h]
\centering
\includegraphics[width=0.33\columnwidth]{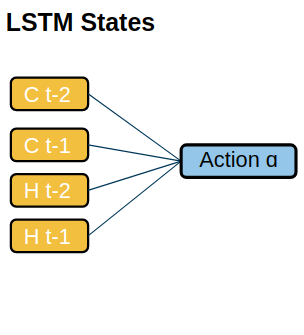}
\caption{The interleaved policy, is based on the LSTM states}
\end{figure}

[3-3.4] The interleaving policy which defines the feature extractor that should be used next is based on reinforcement learning. The state space of the reinforcement learning policy network $\pi$ consists of the LSTM states, $c\textsubscript{t}$, $h\textsubscript{t}$, as well as the differences between the states at different timestamps i.e. $c\textsubscript{t}-c\textsubscript{t-1}$, $h\textsubscript{t}-h\textsubscript{t-1}$ and lastly the action history $\eta\textsubscript{t}$. The action space has length \textit{m} and the action \textit{a} means that the feature extractor f\textsubscript{\textit{a}} should be run. \newline

[3-3.5] It is observed by the authors that, despite the employment of the interleaving policy the real-time detection is limited by the execution of the expensive f\textsubscript{1} feature extractor. They introduce an asynchronous framework for running the feature extractors f\textsubscript{1} and f\textsubscript{2} extractors in parallel. During testing, this asynchronous framework provides better results. \newline

[4-4.1; 4-4.2] As mentioned before, the authors use ImageNet VID 2015 dataset for training and evaluation, which has 30 classes in total, along with the addition of extra data from ImageNet DET and MS COCO17, but this extra data is limited to the classes contained within ImageNet VID. All results are reported, using a Pixel 3 mobile device. The results are compared with, baseline single-frame detection model i.e. MobilenetV2-SSDLite (mAP: 0.420, FPS: 14.4), LSTM-based model i.e. MobilenetV2-SSDLite+LSTM (mAP: 0.451, FPS: 14.6) and the state of the art mobile video object detection model of Zhu et. al. (mAP: 0.602, FPS: 25.6). The proposed model by authors manages to achieve a mAP of 0.593 at 72.3 FPS. The authors also perform evaluation using slight variations of the proposed architecture i.e. Non-interleaved, Interleaved only, Interleaved+Async and Interleaved+Adaptive+Async, in order to test the significance of different components of the architecture on the end results. The Interleaved+Adaptive+Async provides the most balanced end result. \cite{b37}.

\subsection{Detect to Track and track to detect\cite{b8}}

[1] In this paper, Christoph Feichtenhofer, Axel Pinz and Andrew Zisserman, aim to perform, detection and tracking simultaneously using a fully convulational network by infering a "tracklet" over multiple frames.  This paper is based on the R-FCN\cite{b19} model and extends it for multiple frame and tracking. The main contributions of this paper are:
\begin{itemize}
	\item Finding correlation maps of two feature maps of adjacent frames. These correlation maps are used to estimate the local displacement between frames.
	\item Using ROI-tracking layer to regress bounding boxes over multiple frames.
	\item Linking the detections based on tracklets to infer long-term tubes of objects over the course of the video.
\end{itemize}
The proposed model is trained and evaluated on the ImageNet VID dataset \cite{b35}, consisting of 30 classes. The model achieves an overall mAP of 0.82. \newline

\begin{figure}[h]
\includegraphics[width=\columnwidth]{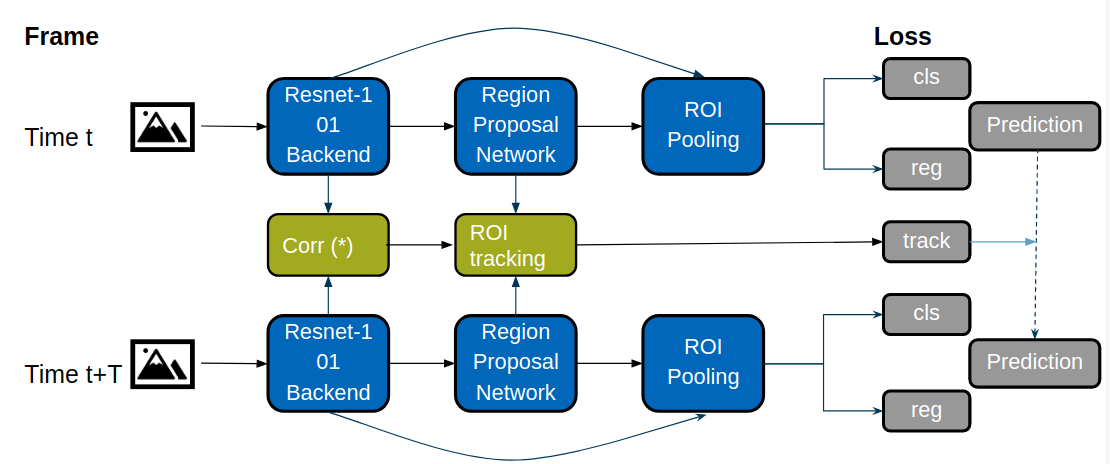}
\caption{Architecture of the model proposed in Detect to Track and track to detect}
\end{figure}

[3-3.2] The R-FCN\cite{b19} detection process consists of two stages, firstly a Region Proposal Network (RPN)\cite{b38} is used to find the candidate region of interests (ROIs) and then a ROI pooling layer\cite{b39} is used to perform classification of the regions into classes or the background. The input to the ROI pooling layer comes from a convulational layer put in place at the end of the ResNet-101 feature extractor\cite{b40} with output $x^t_{cls}$. The ROI layer outputs position sensitive feature maps and using a softmax layer these position senstive feature maps can be converted to class probablities for each ROI. On a second branch, R-FCN places a convulational layer at the end of the ResNet-101 feature extractor which outputs $x^t_{reg}$ and this ouput is again used as an input to ROI pooling layer which generates the bounding boxes. \newline

[3-3.3] The objective function is given hereby:
\begin{equation*}
\begin{aligned}
L(\{p_{i}\},\{b_{i}\},\{\Delta _{i}\}) = \frac{1}{N}\sum_{i=1}^{N}L_{cls}(p_{i,c^{*}})
\\+ \lambda \frac{1}{N_{fg}}\sum_{i=1}^{N}[c_{i}^{*}>0]L_{reg}(b_{i},b_{i}^{*})\\+ \lambda \frac{1}{N_{tra}}\sum_{i=1}^{N_{tra}}L_{tra}(\Delta _{i}^{t+\tau}, \Delta _{i}^{*, t+\tau})
\end{aligned}
\end{equation*}
In short, the first term contains $L_{cls}$, which is the classification, cross-entropy loss, the second term contains $L_{reg}$, which is the bounding boxes regression loss and lastly, $L_{tra}$ in the last term, is the tracking loss across two frames. All of the loss terms are normalized. \newline

[3-3.4] For each pair of adjacent frames I\textsuperscript{t}, I\textsuperscript{t+$\tau$}, a bounding box regression layer is introduced that performs position senstive ROI pooling on the concatentation of bounding box regression features $x^t_{reg}$, $x^{t-1}_{reg}$, which are also stacked with correlation maps, to perform bounding box transformation regression between frames. The correlation maps between the two frames are obtained by finding correlation between the feature maps of the two frames. Finding correlation on all the features in the feature maps will result in an explosion of dimensionality so the correlation maps are only limited to the local neighbors. Like mentioned before, the correlation maps are stacked with the bounding box features maps. \newline

[4] In practical uses, using all frames of the video is not the most efficient way of detection and tracking, as a lot of information between adjacent frames is redundant and also due to GPU memory and computational restrictions, only a certain number of frames can be processed in the GPU at the same time. Due to this, the authors employ a technique similar to action localization\cite{b41} to obtain an optimal path through a video. \newline

[5-5.1;5-5.2;5-5.3] As mentioned above, the proposed model is trained and evaluated on the ImageNet VID dataset. A comparison is made with the R-FCN detector (mAP: 0.742), which the proposed model is based on, the ILSVRC 2015\cite{b42} (mAP: 0.738) winner and the ILSVRC 2016\cite{b43} (mAP: 0.762) winner. DT performs the best with a mAP of 0.82. A comparison between slight variations of the DT model are also evaluated. Firstly, using different feature extractor backbones: ResNet-50 (mAP: 0.767), ResNet-101 (mAP: 0.80) and Inception-v4 (mAP: 0.821)\cite{b44} and secondly using the ResNet-101 backbone but a temporal sampling rate $\tau$ of 10 (mAP: 0.786).

\section{Box-Level-based Video Object Detection}

\begin{figure}[h]
\includegraphics[width=\columnwidth]{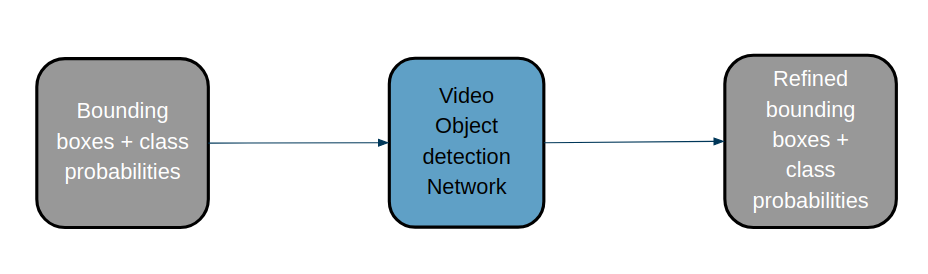}
\caption{Box-Level-based Video Object Detection}
\end{figure}
Bounding Boxes and Class probabilities are fed into the network and are refined temporally and/or spatially.

\subsection{Context Matters: Refining Object Detection in Video with Recurrent Neural Networks \cite{b4}}
In \cite{b4} Tripathi's, Lipton's, Belongie's and Nguyen's architecture consists two parts: A pseudo-labeler, which assigns labels to all video frames and a recurrent unit which refines those pseudo-labels by using the contextual information. Moreover, they describe a training strategy for their architecture and compare their approach to other models on the YouTube-Objects dataset (v2.0) \cite{b25} , which consists of the ten categories airplane, bird, boat, car, cow, dog, horse, mbike, train. Their model reaches an mAP of 68.73 percent which improves the strongest image-based baseline for Youtube-Video Objects dataset of 7.1\% [4-Abstract]. \newline

The final architecture can be found in \textbf{Fig 10}. Tripathi, Lipton and Belongie first train a pretrained YOLO object detection network \cite{b20} as a  pseudo-labeler on the YouTube-Video Dataset. As specified in YOLO  they minimize the weighted squared detection loss and optimize classification and localization error simultaneously [4-3]. 

\begin{figure} [h]
\includegraphics[width=\columnwidth]{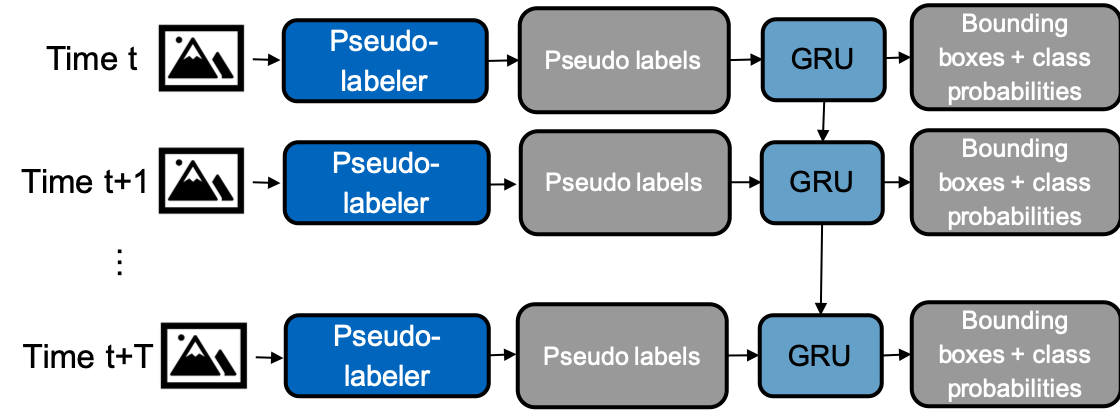}
\caption{Architecture Context Matters: Refining Object Detection in Video with Recurrent Neural Networks \cite{b4}. First the "Pseudolabler" creates bounding boxes and class probabilities for every input frame. Afterwards the GRU (Gated Recurrent Unit) fuses the output of the current and some past frames and refines the bounding boxes and class probabilities.}
\end{figure}

After training the pseudo-labeler they train the Recurrent Neural Network, which takes as an input the pseudo-labels and outputs improved prediction. The RNN consists of two GRU layers [4-2]. \newline

For training the whole network they use the following loss function to take both accuracy at the target frame and consistency of predictions across adjacent time steps into consideration. They choose the values of $\alpha = 0.2$, $\beta = 0.2$ and $\gamma = 0.1$ based on the detection performance on the validation set [4-2.1]: \newline

$ loss = d_{loss} + \alpha \cdot s_{loss} + \beta \cdot c_{loss} + \gamma \cdot pc_{loss} $ \newline

Tripathi, Lipton, Belongie and Nguygen use the object detection loss ($d_{loss}$) as described in YOLO [26, 4-2.1.1]. The similarity loss ($s_{loss}$) considers the dissimilarity between the pseudo-labels and prediction at each frame t [4-2.1.2], the category loss ($c_{loss}$) takes wrong class probabilities into consideration [4-2.1.3] and the prediction-consistency loss ($pc_{loss}$) regularizes the model by encouraging smoothness predictions across the time-steps [4-2.1.4]. \newline

During the evaluation they find two possible areas of improvement for their approach. On the one hand the RNN is not able to recover from wrong predictions made by the pseudo-labeler after they have been fed into the RNN. This is a general disadvantage of box-level methods in comparison to feature-level methods. On the other hand, they observe that their network is not robust to motion [4-3.4]. \newline

Tripathi, Lipton, Belongie and Nguyen test their model on the Youtube-Objects dataset. Overall they outperform the best no-recurrent architecture (DA Yolo) in their comparison by 7\% mAP [4-3.1]. This again confirms the observatoin that adding recurrency to integrate temporal context improves the detection quality.  \newline

The architecture mentioned in "Context Matters: Refining Object Detection in Video with Recurrent Neural Networks" is pretty similar to the one used in \cite{b4}. Both are using the YOLO network architecture as a baseline and feed its output into an Recurrent Unit. Main Differences is that \cite{b4} feeds the bounding boxes and in addition some visual features into the recurrent unit.  

\subsection{Optimizing Video Object Detection via a Scale-Time Lattice\cite{b10}}

[1] The aim of this paper, is to propose an architecture which is balanced and flexible enough to allow prioritisation of accuracy or performance with minimal effort. The primary contributions of this paper published by Kai Chen et. al. are:
\begin{itemize}
	\item The Scale-Time Lattice, which provides a rich design space.
	\item A detection framework which provides good accuracy to performance trade-off.
	\item A novel key-frame extraction policy, which is based on the ease of detection.
\end{itemize}

\begin{figure}[h]
\centering
\includegraphics[width=\columnwidth]{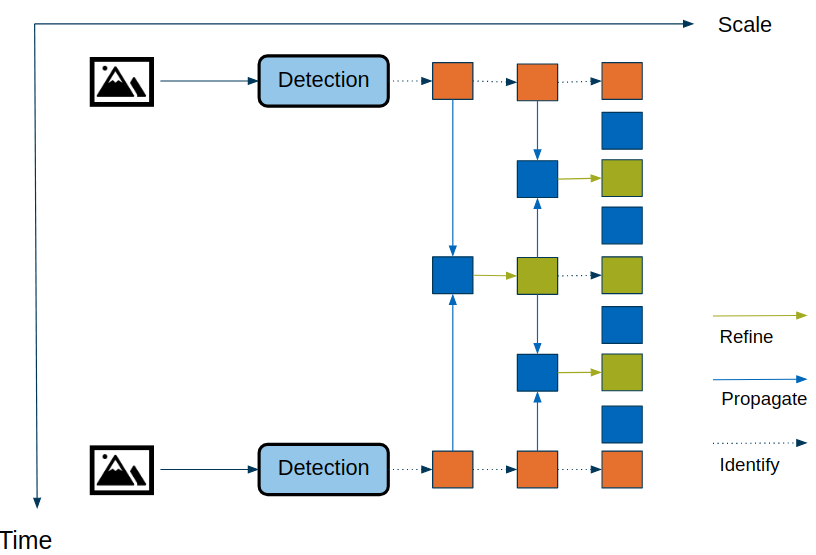}
\caption{The Scale-Time Lattice has an efficient design for reaching a balance between performance and accuracy}
\end{figure}

[3] The Scale-Time Lattice allows coarse detection, both temporally and spatially and then use temporal propagation and spatial refinement to go from coarse to fine detection. \newline

\begin{figure}[h]
\centering
\includegraphics[width=0.5\columnwidth]{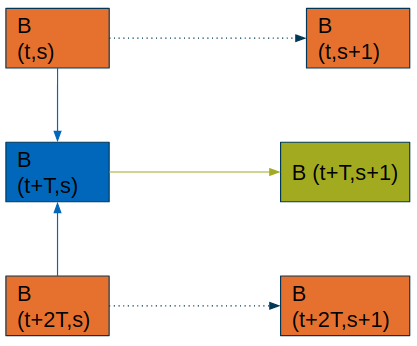}
\caption{The Propagation and Refinement Units (PRUs) are the constituent elements which the Scale-Time Lattice is made up of}
\end{figure}

[4-4.1] The Scale-time lattice is composed of structures which connect together to perform the temporal and spatial operations. These structures are called Propagation and Refinement Units (PRUs). PRUs work on the basis of two operators $F_\tau$ and $F_S$, and by carefully allocating resources to the two operators, a balance between the detection accuracy and the performance can be acheived. Assuming that two frames are sampled from a video at time $t$ and $t+2\tau$, both at scale $s$, the $F_\tau$ operator tries to model the movement of the bounding boxes from time $t$ to $t+\tau$ and from time $t+2\tau$ to $t+\tau$ irrespective of the scale offset between the scale $s$ and the ground truth frame scale $s+1$. The modeling of the bounding boxes offset from scale $s$ to $s+1$ is the task of the $F_S$ operator. The figure depicting the PRU gives more information on this. \newline

[4-4.1] The objective function is given hereby:
\begin{equation*}
\begin{aligned}
L(\Delta_{F_{\tau}} ,\Delta_{F_{S}} ,\Delta_{F_{\tau}}^{*} ,\Delta_{F_{S}}^{*} ) = \\ \frac{1}{N}\sum_{j=1}^{N}L_{F_{\tau}}(\Delta_{F_{\tau}}^{j} ,\Delta_{F_{\tau}}^{*j}) + \\ \lambda \frac{1}{N}\sum_{j=1}^{N}L_{F_{S}}(\Delta_{F_{S}}^{j} ,\Delta_{F_{S}}^{*j})
\end{aligned}
\end{equation*}
Here, N is the number of bounding boxes in the batch, while $\Delta_{F_{T}}$ and $\Delta_{F_{S}}$ are the network output of $F_{\tau}$ and $F_{S}$. The $L_{F_{\tau}}$ and $L_{F_{S}}$ are losses of temporal propagation and spatial refinement network, respectively. \newline

[4-4.2] The most straight-forward keyframe extraction policy is a uniform i.e. to use keyframes from the video after uniform intervals but an intelligent keyframe extraction policy can be used to increase the accuracy of the results. To that effect, the authors introduce a ease of detection cofficient e. If during detection, the value of e falls beneath a certain threshold that the sample rate of the frames from the video is increased. This can happen if there are a lot of objects on the screen or the objects are moving too quickly. \newline

[5-5.2] The proposed model has an mAP of 0.79 at 60 FPS on the ImageNet VID dataset. \newline

\subsection{Spatially Supervised Recurrent Convolutional Neural Networks for Visual Object Tracking \cite{b5}}
Ning, Zhang, Huang, He, Ren and Wang design in \cite{b5} a combination of box-level and feature-level based Video detectors. They use the YOLO\cite{b26} network to create high-level visual features and preliminary location inferences and feed both into a recurrent unit. They test their approach on the OTB-30 dataset \cite{b28} and compared it with 9 different state-of-the-art trackers and outperform all of them. Their architecture, called ROLO, is shown in figure 11. [5-Abstract; 5-4] 

\begin{figure} [h]
\includegraphics[width=\columnwidth]{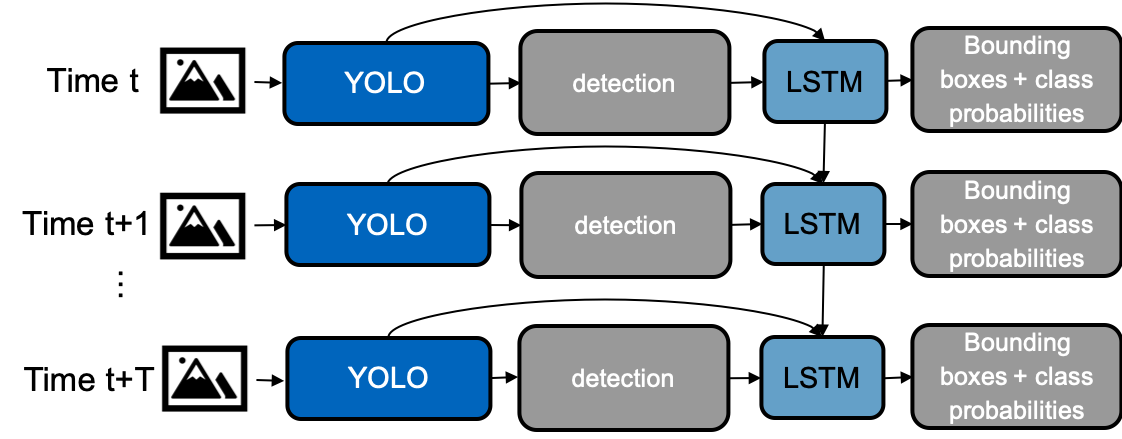}
\caption{Architecture ROLO \cite{b5}. The architecture feeds feature representations made by the YOLO network as well as bounding boxes also made by the YOLO Network into a recurrent unit. The recurrent unit uses the temporal context of those inputs to finally output bounding boxes and class probabilities.}
\end{figure}

The architecture consists two main parts. The YOLO\cite{b26} networks which collects visual features and outputs also preliminary location inferences and the LSTM which is used as a tracking module [5-2]. \newline 

They use three phases to train their model: The pre-training phase of  convolutional layers for feature learning, the traditional YOLO training phase for object proposal and the LSTM training phase for object tracking. [5-3]  \newline

In the pre-training phase convolutional layers, which create feature maps are pretrained on ImageNet data. Afterwards the YOLO-architecture is adopted as detection module and trained by using the traditional YOLO training phase. Lastly they add LSTMs for tracking. The LSTMs are fed with, the feature representations from the convolutional layers, the Bounding boxes from the detection module and the output states from the LSTM in the previous time-step. For training they use the Mean Squared Error (MSE) [5-3]: \newline

\[ L_{MSE} = \dfrac{1}{n} \sum_{i = 1}^n ||B_{target}-B_{pred}||_{2}^2 \] \newline

As an alternative, Ning, Zhang, Huang, He, Ren and Wang mention a Heatmap as  input for the LSTM. Therefor the prediction and the visual features are concatenated before adding them to the LSTM. This is helpful to visualize the intermediate results. [5-3.3] \newline

Special to this architecture is, that LSTM does regression in two folds. There is a regression within one frame between the location inferences and the high-level features and there is also regression over the different frames of the sequence by taking the temporal context into account [5-3.4].  \newline

In 2015 Wu, Lim and Yang published a comparison of different online object trackers on their one data set \cite{b28}. The best 9 architectures (STRUCK [47]; CXT [48]; OAB [49]; CSK [50]; VTD [51]; VTS [52]; LSK [53]; TLD [54]; RS [55]) were used by Ning, Zhang, Huang, He, Ren and Wang to compare them with their own architecture. They use 30 videos with different objects (e.g.: car, human, Skater, Couple,..) out of the benchmark \cite{b28} dataset. They get the best tracking result in 13 videos and the second best in 3 videos. There is no other tracker that reaches the best results in more then 4 videos. \newline

This approach combines box-level and feature-level methods and reaches good results based on the combination of spatial regression between features and location proposals and the temporal regression in the LSTM. The paper shows a potential way to combine the best of both methods (feature based and box-level approaches) which are shown in the other papers. 

\section{Flow-based Object Detection}

\subsection{Definition}
Another type  of architectures for Video Object Detection defined, for example in \cite{b3}, are architectures which use Flow-Networks to consider the temporal context. The flow network estimates the optical flow which means it projects back the location in the current frame to an earlier frame. 

\subsection{Deep Feature Flow for Video Recognition \cite{b3}}
Zhu, Xiong, Dai, Yuan and Wei develop in \cite{b2} a way to use flow nets to detect objects in video frames. As shown in figure 12 their architecture consists of three main parts: A network to create visual features, a network to create class probabilities and bounding boxes respectively the segmentation out of feature maps, and a network which estimates the optical flow. They tested their approach on the Cityscapes dataset \cite{b56}  with scenes for autonomous driving and ground-truth for semantic segmentation and the ImageNet VID dataset \cite{b35} with ground truth for object detection.

\begin{figure} [h]
\includegraphics[width=\columnwidth]{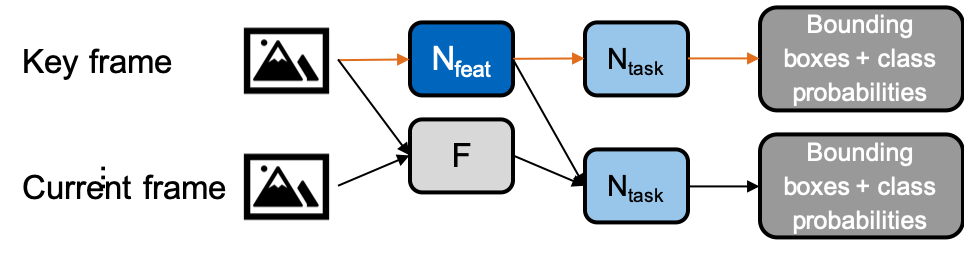}
\caption{Architecture Deep Feature Flow for Video Recognition \cite{b3}. The feature Network $N_{feat}$ extracts visual features for the key frames. The $N_{task}$ networks uses visual features to create Bounding boxes and Class probabilities. For non-key frame images, the flow networks F estimates the feature maps based on in a image sequence.} 
\end{figure}

The paper differs between key frames and other frames. The network to create visual features $ N_{feat} $ only processes on key frames. It is a fully convolutional network, which takes as an input the image and outputs a feature maps. As a feature network they use a pretrained version of ResNet \cite{b40}. [3-3, 3-4] \newline

The second network $ N_{task} $ does the recognition task over the feature maps. It performs on every frame.  Zhu, Xiong, Dai, Yuan and Wei use an R-FCN \cite{b19} network for the recognition task. [3-3, 3-4]  \newline

For all non-key frames the feature maps are not constructed by the feature network. Instead, they are propagated by a flow network $ N_{flow} $.  Zhu, Xiong, Dai, Yuan and Wei use a state-of-the-art CNN based FlowNet architecture \cite{b57} as a flow network. The flow estimation is given by [3-3, 3-4]: \newline

$ f_{i} =  W (f_{k},M_{i->k}, S_{i->k})$ \newline

$ f_{k} $ is the key frame's feature map, $ M_{i->k} $ is a two dimensional flow field, it projects back the location of an object in the current frame to the location in the key frame by using bilinear interpolation and  $ S_{i->k} $ is scale field [3-3]. \newline

Zhu, Xiong, Dai, Yuan and Wei use a fixed key frame scheduling and they  see a potential improvement in changing the key frame policy. \newline

Overall Zhu, Xiong, Dai und Yuan show  a alternative  to recurrent neural networks to integrate the temporal context into the object tracking. They reach an mIoU up to 69.2 percent on Cityspace dataset and an mAP of 73.9 percent on ImageNet Vid. But those results are reached by processing with 5.6 and 4.05 fps. Both drop significantly with higher fps. Some recurrent papers in 2 and 3 reach higher mAPs on higher fps. 

\section{Comparison of different approaches}

\subsection{KITTI Dataset}
\captionof{table}{Results on KITTI Dataset} 
\begin{tabular}{ | p{2cm} | p{2em}| p{2em} | p{4em} | p{5em} | } 
 \hline
 Model & MAP & FPS & Machine & Architecture \\
 \hline
 Recurrent \cite{b1} & 86.0 & 50 & Nvidia TITAN X & Feature-Level \\
 \hline
 Feature Selective \cite{b6} & 81.3 & 30.8 & Nvidia TITAN X & Feature-Level \\
 \hline
\end{tabular} \newline

As seen in Table I the architecture mentioned in "Recurrent Multi-frame Single Shot Detector for Video Object Detection" \cite{b1} outperforms the results of the paper "Feature Selective Small Object Detection via Knowledge-based Recurrent Attentive Neural Network" \cite{b6} in regard to the detection quality (mAP) and the computational speed (fps). \newline

The main difference between those two architectures is that \cite{b6} performs on a single frame and uses LSTMs to include the spatial context within this frame. \cite{b1} instead is processing on a sequence of input frames and uses recurrent units to take the temporal context into consideration. \newline

That leads to the hypothesis that performing on multiple frames is more beneficial than performing on only one frame. Which means that temporal context is more important than spatial context.   

\subsection{ImageNet Dataset}

\captionof{table}{Results on ImageNet Dataset} 
\begin{tabular}{ | p{2cm} | p{2em}| p{2em} | p{4em} | p{5em} | } 
 \hline
 Model & MAP & FPS & Machine & Architecture \\
 \hline
 DT \cite{b8} & 82.0 & 7 & Nvidia TITAN X & Feature-Level \\
 \hline
 DT \cite{b8} & 78.5 & 55 & Nvidia TITAN X & Feature-Level \\
 \hline
 Scale-Time Lattice \cite{b10} & 79.6 & 20 & Nvidia TITAN X & Box-Level \\
 \hline
 Scale-Time Lattice \cite{b10} & 79 & 62 & Nvidia TITAN X & Box-Level \\
 \hline
 DeepFeature Flow \cite{b3} & 73.9 & 3 & - & Flow-Based \\
 \hline
 DeepFeature Flow \cite{b3} & 73.1 & 20.5 & - & Flow-Based \\
 \hline
 Looking Fast and Slow \cite{b7} & 60.7 & 48.8 & Pixel Phone & Feature-Level \\
 \hline
 Object Detection with Temporally-Aware \cite{b2} & 54.4 & 15 & Pixel Phone & Feature-Level \\
 \hline
\end{tabular} \newline

As seen in Table 2 both "Detect to Track and Track to Detect" and "Optimizing Video Detection with spatial-time lattice" reach better results than the other papers. Both "Detect to Track and track to Detect" and "Optimizing Video Detection with spatial-time lattice" perform on multiple frames in parallel. That leads to the advantage that both of them can use temporal context from the past and also from the future. The doubling of the temporal context information is probably one reason for their comparatively good performance. \newline

In addition to the advantage of performing on multiple frames in parallel "Optimizing Video Detection with spatial-time lattice" uses not only the temporal context. In addition to the temporal context, this approach takes also different scales into consideration. This is a possible further reason for the good results. \newline

The Flow-based approach \cite{b3} has comparatively bad results on ImageNet. We only evaluated one flow-based paper, but the comparatively bad performance could be an evidence that flow-based approaches' benefit in comparison to recurrent approaches does not exist or is very small. \newline

"Looking Fast and Slow: Memory-Guided Mobile Video Object Detection" and "Mobile Video Object Detection with Temporally-Aware Feature Maps" cannot be compared to the other 3 approaches because they are running on mobile devices, which have pretty less computational power than the GPUs which are used by the other papers.  

\subsection{Results on COCO Dataset}

\captionof{table}{Results on COCO Dataset} 
\begin{tabular}{ | p{2cm} | p{2em}| p{2em} | p{4em} | p{5em} | } 
 \hline
 Model & MAP & FPS & Machine & Architecture \\
 \hline
 Feature Selective \cite{b6} & 57.8 & 37.5 & Nvidia TITAN X & Feature-Level \\
 \hline
\end{tabular}

\subsection{Results on YouTube Dataset}

\captionof{table}{Results on YT Dataset} 
\begin{tabular}{ | p{2cm} | p{2em}| p{2em} | p{4em} | p{5em} | } 
 \hline
 Model & MAP & FPS & Machine & Architecture \\
 \hline
 Context Matters \cite{b4} & 68.73 & - & - & Box-Level \\
 \hline
\end{tabular} \newline

"Context Matters" is the only paper of those which we have done some research on, which uses the YouTube Dataset to test their architecture. Unfortunately the results cannot be directly compared to the other papers. Also Tripathi, Lipton, Belongie and Nguygen only compare their model with non-recurrent ones. \newline

\subsection{Results on OTB Challenge Dataset}
\captionof{table}{Results on OTB Challenge Dataset} 
\begin{tabular}{ | p{2cm} | p{3em}| p{2em} | p{4em} | p{4em} | } 
 \hline
 Model & Success Rate & IoU & FPS & Machine \\
 \hline
 Spatially Supervised \cite{b5} & 0.564 & 0.455 & 20/60 & Nvidia TITAN X \\
 \hline
\end{tabular} \newline

Out of the paper which are mentioned in this paper only "Spatially Supervised Recurrent Convolutional Neural Networks for Visual Object Tracking" uses the OTB Challenge Dataset to evaluate their results. Unfortunately, they are only doing a comparison with non-recurrent approaches. 

\section{Outro}

\subsection{Conclusion}
The main conclusion of our research is that temporal context matters. All the papers came to the results that their models, which use more than one frame to detect objects outperform the models with similar baseline architecture which only proceed on single frames. \newline

Moreover, we have seen that operating on multiple-frames at the same time is a beneficial approach. It doubles the amount of temporal context information, which leads to higher mAPs. \newline

With respect to the computational speed we noticed that the recurrent units should not be too deep. And in addition working only on some keyframes can be beneficial to increase the speed. Therefor a good key-frame policy is needed. \newline

For detection quality and also for computational speed it is beneficial to work on different scales. This enables us to use recurrency to take the spatial and the temporal context into consideration. 

\subsection{Further work}
\begin{figure} [h]
\includegraphics[width=\columnwidth]{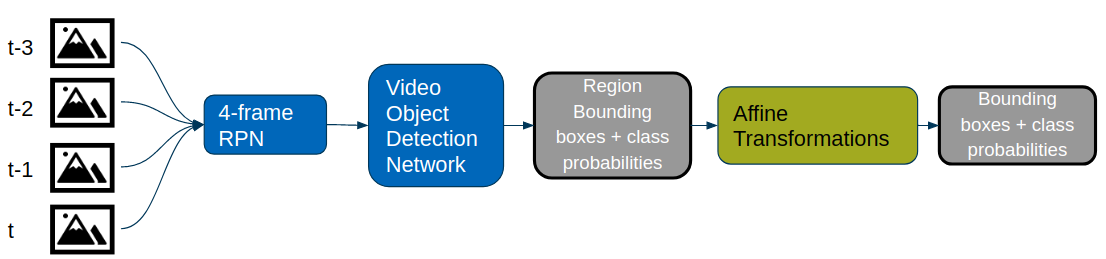}
\caption{The proposed architecture} 
\end{figure}

The proposed architecture consists of a region proposal network \cite{b38} that is based on the N-Gram concept in Natural Language Processing. Given a window of N previous frames, the RPN proposes the regions where the object bounding boxes could be detected from within the next frame. The RPN (region proposal network) should be recurrent in nature for detecting the temporal dependencies and it is to be really light weight. The ROIs obtained from RPN are fed into the network to make the predictions. So rather than feeding the whole image, feed only the region proposals made by RPN. Lastly, affine transformations can be performed to the output bounding boxes to overlay them over the image.

\end{document}